\documentclass[letterpaper]{article} 
\usepackage{aaai24}  
\usepackage{times}  
\usepackage{helvet}  
\usepackage{courier}  
\usepackage[hyphens]{url}  
\usepackage{graphicx} 
\usepackage{natbib}  
\usepackage{caption} 
\usepackage{amsmath,amsfonts}
\usepackage{algorithm}
\usepackage{algorithmic}

\usepackage{newfloat}
\usepackage{listings}
\DeclareCaptionStyle{ruled}{labelfont=normalfont,labelsep=colon,strut=off} 
\floatstyle{ruled}
\newfloat{listing}{tb}{lst}{}
\floatname{listing}{Listing}
\nocopyright
\usepackage{multirow, multicol}
\usepackage{longtable, threeparttable, booktabs}
\usepackage{amsmath}
\usepackage{subfigure}

\title{Ranking-aware Uncertainty for Text-guided Image Retrieval}

\author {
    Junyang Chen\textsuperscript{\rm 1},
    Hanjiang Lai\textsuperscript{\rm 1}\thanks{Corresponding author}
}
\affiliations {
    \textsuperscript{\rm 1} School of Computer Science and Engineering, Sun Yat-Sen University\\
     chenjy855@mail2.sysu.edu.cn, laihanj3@mail.sysu.edu.cn
}

\usepackage{bibentry}

\begin{document}

\maketitle

\begin{abstract}
Text-guided image retrieval is to incorporate conditional text to better capture users' intent. Traditionally, the existing methods focus on minimizing the embedding distances between the source inputs and the targeted image, using the provided triplets $\langle$source image, source text, target image$\rangle$. However, such triplet optimization may limit  the learned retrieval model to capture more detailed ranking information, e.g., the triplets are one-to-one correspondences and they fail to account for many-to-many correspondences arising from semantic diversity in feedback languages and images. To capture more ranking information, we propose a novel ranking-aware uncertainty approach to model many-to-many correspondences by only using the provided triplets. We introduce uncertainty learning to learn the stochastic ranking list of features. Specifically, our approach mainly comprises three components: (1) In-sample uncertainty, which aims to capture semantic diversity using a Gaussian distribution derived from both combined and target features; (2) Cross-sample uncertainty, which further mines the ranking information from other samples' distributions; and (3) Distribution regularization, which aligns the distributional representations of  source inputs and targeted image. Compared to the existing state-of-the-art methods, our proposed method achieves significant results on two public datasets for composed image retrieval.
\end{abstract}

\section{Introduction}

Text-guided image retrieval (TGIR)~\cite{vo2019TGIR}, which aims to retrieve the image that better matches the user's intent by integrating the reference image and text feedback as a query, has received a lot of attention. Compared with the traditional image-only modal retrieval, the combination of textual modality enables users to express their thoughts more flexibly. TGIR can improve the user experience for search, which is more in line with the user's needs.  

Recently, considerable research effort~\cite{TIRG, Chen_2020_CVPR, zhang2020joint, CLVC-Net, CLIP4Cir, chen2022composed} has been devoted to text-guided image retrieval.
The training of these works is performed with triplets $\langle$source image, source text, target image$\rangle$ provided by TGIR dataset~\cite{guo2019fashion, liu2021image}. Hence, most of the previous work has been directed towards a multi-modal similarity metric approach, i.e., how to reasonably fuse the features of the source image and source text, and the combined feature is highly similar to the feature of the target image. For example, CLIP4Cir~\cite{CLIP4Cir} learned the multi-modal similarity metric based on CLIP~\cite{Radford2021CLIP} features. \cite{CLVC-Net} proposed local-wise and global-wise compositions for TGIR. Another research approach~\cite{yan2020deep, Warburg2021iccv, chen2022composed} aims to improve the generalisation ability of the retrieval model via data enhancement. For instance, \cite{chen2022composed} introduced a simple Gaussian noise to the target image features. 

\begin{figure}
    \centering
    \includegraphics[width=\linewidth]{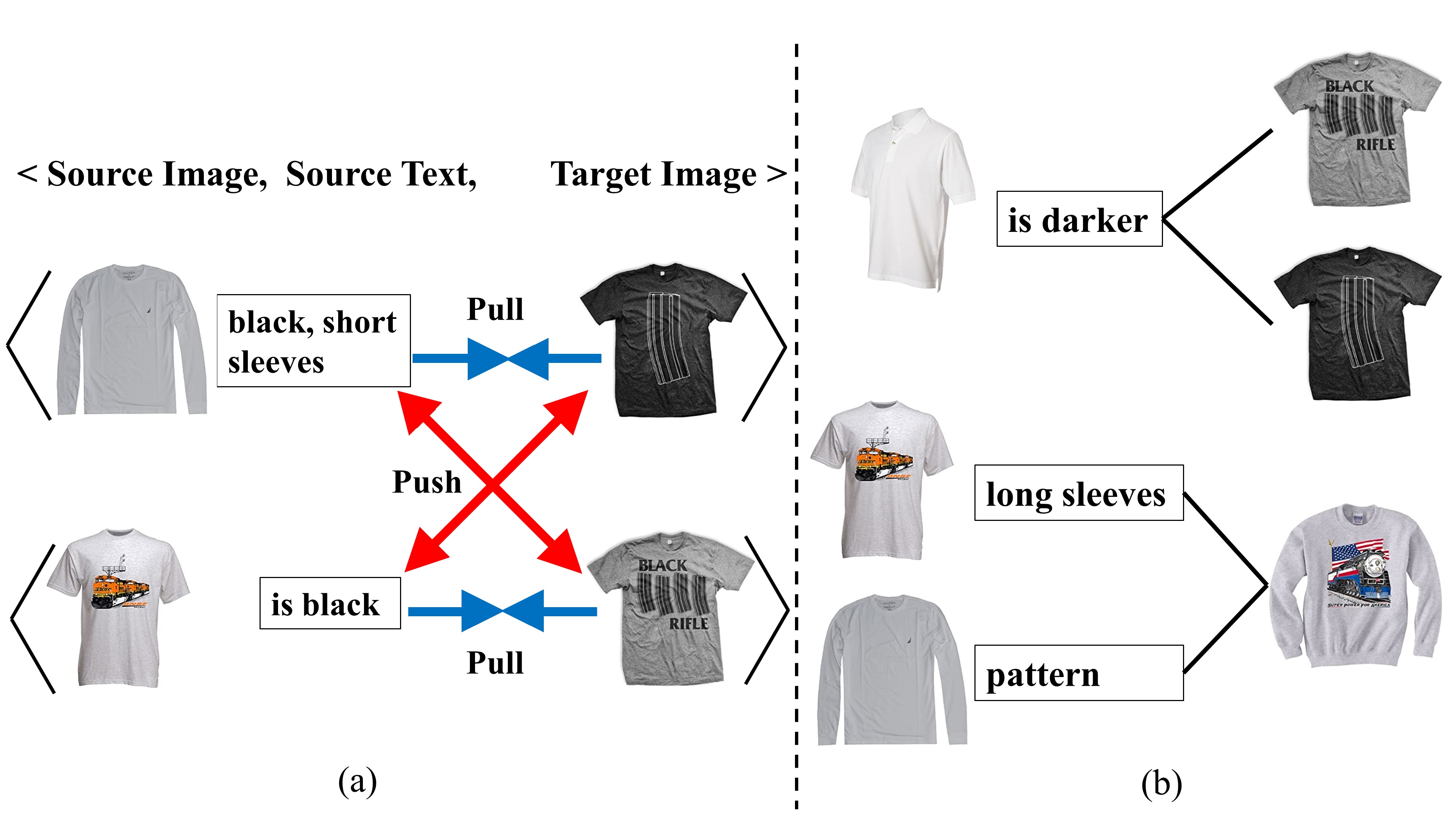}
    \caption{(a) Triplet optimization for the existing methods. Only the source input or target image in the same triplet is the positive sample, others are negative. (b) Many-to-many ranking-aware optimization. Due to semantic diversity, we further consider the many-to-many correspondences only using the provided triplets in this paper.}
    \label{fig:problem}
\end{figure}
However, such triplet optimization only considers the one-to-one correspondences but ignores that the retrieval model should rank a list of samples according to their relevances. As shown in Figure \ref{fig:problem}  (a), the triplet optimization only moves the target' feature close to the combined feature of the source image and text, and others are considered as the negative samples and be pushed away. Such triplet optimization may hurt the performance of the learned retrieval model due to the semantic diversity in languages and images. For example, in Figure~\ref{fig:problem}  (b), ``Darker color" for  white cloth can be either gray or black cloth.  Different source images and texts can have the same target image. Unfortunately, most of the previous work only considers one-to-one triplet optimization and ignores the fact that TGIR is a ranking task with many-to-many correspondences. 

In this paper, we propose a novel ranking-aware uncertainty approach to unlock the limitations of triplet optimization. This algorithm addresses the one-to-one correspondences by using stochastic mapping instead of deterministic mapping to formulate the many-to-many correspondences. That is image/text is not encoded to a deterministic feature but a distribution of feature space. In this paper, we propose three main components for exploring more ranking information: in-sample uncertainty, cross-sample uncertainty, and distribution regularization. The in-sample uncertainly module extends the provided triplets and simulates a many-to-many situation. It not only considers the similarity in the original triplet (e.g., the target image) but also the other similarities  (e.g., generated from the distributions). Then, we further explore the ranking similarities from other samples' distributional spaces by the cross-sample uncertainty module. Finally, distribution regularization is proposed to align the combined and target distributions. Experimental results have indicated that our proposed ranking-aware method performs significantly better than the existing state-of-the-art baselines. Our method achieves 42.50\% of R@10, which indicates an increase of 9.33\% compared to the second-best baseline. 

 The main contributions of our work are summarized as:
 \begin{itemize}
     \item We reformulate the text-guided image retrieval task, which considers not only the one-to-one triplet optimization but also the many-to-many ranking-aware optimization. 

     \item We propose a novel ranking-aware uncertainty for TGIR, which can explore the ranking-aware optimization without the additional manual labeling. 

     \item Extensive experimental results demonstrate the compelling performance of our method compared to the SOTA baselines.
 \end{itemize}

\section{Related Work}

\textbf{Text-guided Image Retrieval.} Previous work has focused on how to appropriately combine two modal inputs for TGIR. Several works~\cite {Chen_2020_CVPR, Hosseinzadeh2020locally, zhang2020joint} had proposed to combine the modified textual representations with local visual descriptors of source images to query target image representations. Instead, TIRG~\cite{TIRG} implemented a modification of the global representation of the source image that encourages cross-modal feature learning with gating and residual designs. MAFF~\cite{dodds2020modality_maaf} fused modality-agnostic features obtained from spatial convolutional layers and LSTM hidden states. DCNet~\cite{kim:2021:AAAI} simply cascaded global and local features to obtain a more robust representation of the source image. Further, CLVC-Net~\cite{CLVC-Net} was designed with two split sub-networks that mutually enhance each other by sharing knowledge with each other during the alternative optimization process to achieve fine-grained local and global combinations, respectively. CLIP4Cir~\cite{Baldrati2022combiner} extracts text and image features using the prowess of pre-trained CLIP models and designs a non-linear combiner for feature fusion. However, these existing works only considered the one-to-one triplet optimization and our method extends the triplet optimization to ranking optimization by exploring more many-to-many correspondences. 

\textbf{Uncertainty Learning.} Uncertainty is used as a measure of the ``confidence" in a prediction, i.e., how reliable the model is. In general, uncertainty can be divided into model (epistemic) uncertainty and data (aleatoric) uncertainty~\cite{Kendall2017nips}. Model uncertainty means that the model's estimate of the input data may be inaccurate due to poor training, insufficient training data, independent of a single piece of data, etc. For example, Bayesian neural networks~\cite{Parsons2008bn, Gal2016bnn} modeled the inherent uncertainty of individual parameters by learning the probability distribution of the weights. Monte Carlo Dropout~\cite{Gal2016bnn} simulated a Bayesian network by dropping some neurons randomly. Data uncertainty describes the noise inherent in the data, such as the ambiguity of labeled data. \cite{HeZWS019cvpr} proposed a KL loss to learn the bounding box transform and localization variance for the problem of fuzzy labeled boundaries in target detection datasets, thus improving the detection accuracy without increasing the number of parameters. \cite{chen2022composed} has modeled coarse-grained matching in TGIR by introducing Gaussian noise modeling uncertainty in the feature space. Different to \cite{chen2022composed}, it only modeled one-to-many correspondences and our method can model many-to-many correspondences. 

\section{Method}

\begin{figure*}
    \centering
    \includegraphics[width=0.9\linewidth]{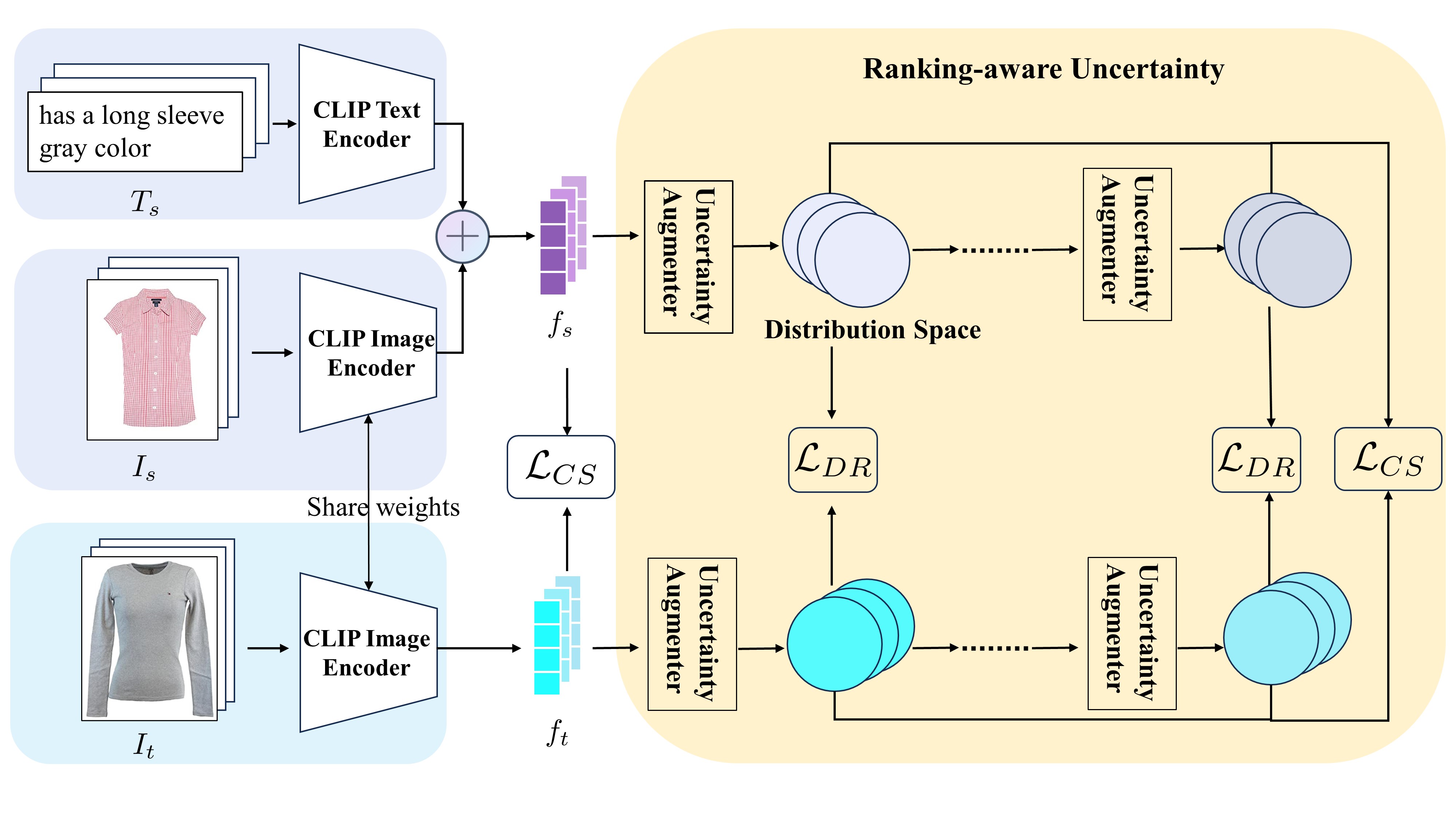}
    \caption{Overview of the proposed method. Given a batch of triples $\langle$source images, source texts, target images$\rangle$ denoted by $\langle I_s, T_s, I_t \rangle$, we extract the features by Clip's encoder and get the additive combined features $f_s$ and the target features $f_t$ respectively. Then a many-to-many relationship is constructed on this batch of features using Ranking-aware Uncertainty. Notice that Ranking-aware Uncertainty is a plug-and-play method and is only used to train the model.}
    \label{fig:overview}
\end{figure*}

In this section, we first give the problem formulation for the text-guided image retrieval problem.  Also, the existing triplet optimization loss objective will be simply introduced. Then, we will present our proposed ranking-aware uncertainty method for TGIR. 

\subsection{Problem Formulation}
In the text-guided image retrieval task, an image and a text are used as a query to retrieve the desired image. To learn such retrieval models, a large number of triples, i.e., $\langle$source image, source text, target image$\rangle$, are provided. We denote the source image, source text, and target images by $I_s, T_s, I_t$, respectively. Given many triplets in the training dataset, we aim to learn two embedding functions $f_s = F_s(I_s,T_s)$ and $f_t=F_t(I_t)$, where $F_s$ takes the source image and text as input and obtains their combined feature, and $F_t$ map the target image into a feature representation. Many methods have been proposed to use deep neural networks for embedding functions $F_s$ and $F_t$. For example, the images and text are encoded into features using their respective CLIP encoders. For each triplet  $\langle I_s, T_s, I_t \rangle$, the learned $f_s$ should be similar to $f_t$.

For triplet optimization, contrastive loss (CL)~\cite{chen2020simcl} is often used as a ranking loss objective in many existing methods~\cite{TIRG, 2021CoSMo, CLIP4Cir} for TGIR, which can be formulated as batch-based similarity loss:
\begin{equation}
        L_{CL}{ ( f_ s , f_ t ) }=\frac1B\sum_{i=1}^B-\log\frac{\exp\left(S\left(f_s^i,f_t^i\right)\right)}{\sum_{j=1}^B\exp\left(S\left(f_s^i,f_t^j\right)\right)},
\end{equation}
where $S(\cdot)$ is cosine similarity and $B$ is mini-batch size. In the $\mathcal{L}_{CL}$, only the entry from the same triplet is positive and other entries within the batch as treated as negative samples. Such optimization may ``confuse" the  embedding functions. For example, given a triplet, e.g., a white T-shirt + black $\to$ a black T-shirt, if we have another black T-shirt in the mini-batch, there will be a conflict in triplet optimization: the white T-shirt + black is moved close to the black T-shirt but also moved away from the black T-shirt. It is a conflict and the performance also would be degraded.

\subsection{Overview}

To address this issue, the outline of the proposed method is shown in Figure~\ref{fig:overview}. In our method, we do not need any additional manual labeling, and also a batch of triples are taken as inputs. Similar to the existing works, these inputs are firstly encoded into features in the common space using deep neural networks. Then, the text and source image features are simply fused, and finally, the source feature $f_s$ and target feature $f_t$ are obtained. Since our target is not on the deep neural networks and fusion, we simply use the CLIP encoders to extract the features and simply concatenate the features of the source image and text to obtain $f_s$.  

To model many-to-many correspondences, we propose a ranking-aware uncertainty learning, which further maps the source and target point features into distributions. More specially, the proposed method mainly includes three modules to explore the many-to-many ranking optimization: in-sample uncertainty, cross-sample uncertainty, and distribution regularization. In the following, we will present the details of these three modules.

\subsection{In-Sample Uncertainty}
\textbf{Uncertainty augmenter:} Now we have point features $f_s$ and $f_t$, we introduce an uncertainty augmenter (UA) to model many-to-many correspondences. 
The UA expresses richer semantic relationships by learning distributions instead of point features. Please note that we can get multiple instances by sampling from a distribution. Thus it can learn many-to-many mapping relationships only using the one-to-one triplets. 

\begin{figure}
    \centering
    \includegraphics[width=\linewidth]{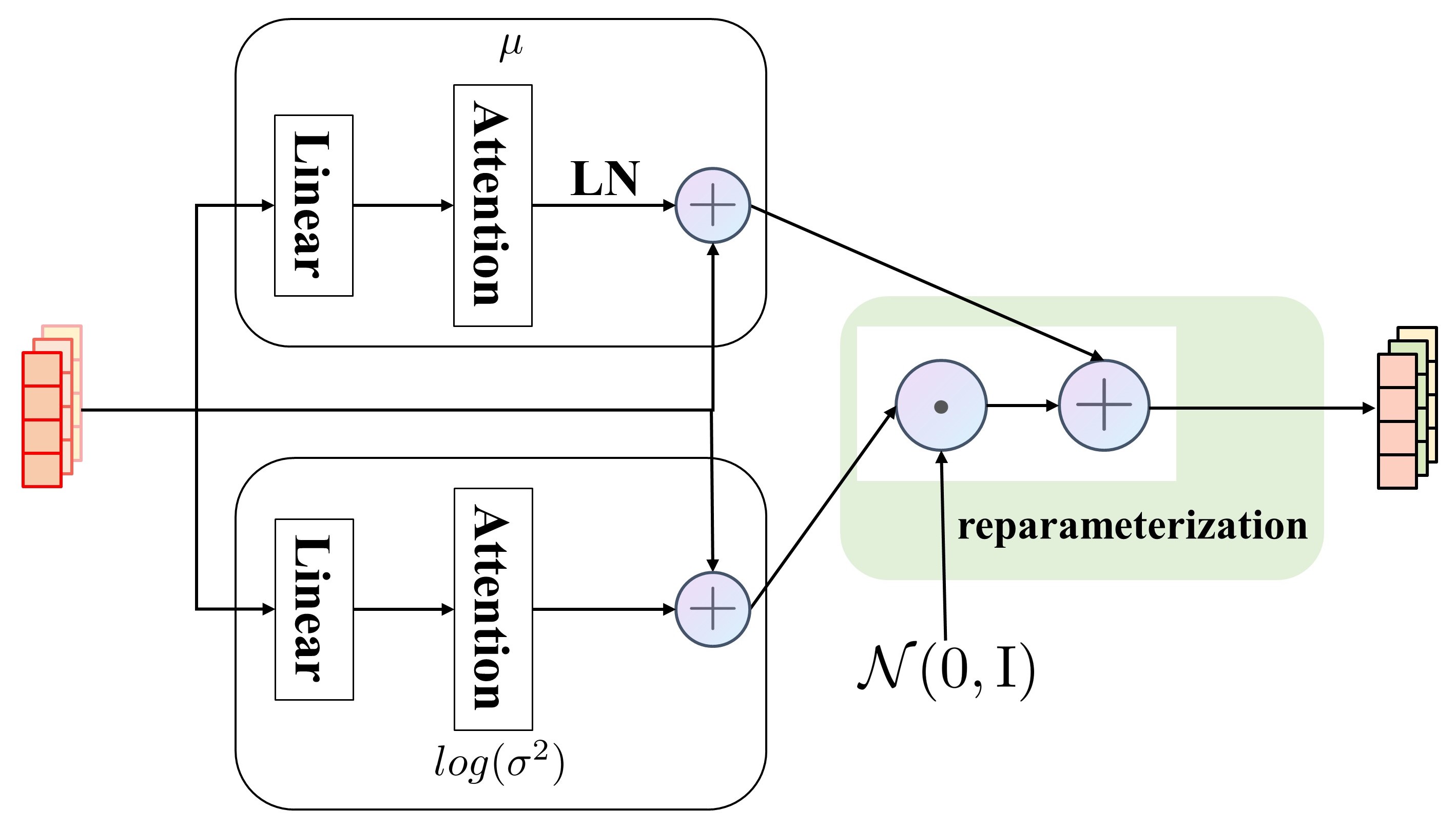}
    \caption{The architecture of uncertainty augmenter (UA) block.}
    \label{fig:ua}
\end{figure}

To model these distributions, we simply frame the input features as multivariate Gaussian distributions. As shown in Figure \ref{fig:ua}, given an input $f$, UA outputs a mean vector and a variance vector for the input. In particular, the variance vector expresses the uncertainty of the samples through fluctuations. Specifically, inspired by~\cite{chun2021probabilistic, neculai2022probabilistic}, for the input feature $f$, the UA models it as a Gaussian distribution $\mathcal{N}\left( \mu, \Sigma \right)$, where $\Sigma$ is the diagonal variance matrix. The specific calculation procedure is as follows:
\begin{equation}
    \begin{aligned}
    &\mu=\mathrm{LN}((f+\mathrm{fc}(\mathrm{attn}(f))),\\
    &\log(\sigma^2)=f+\mathrm{fc}(\mathrm{attn}(f)),
    \end{aligned}
\end{equation}
where $\sigma$ refers to a standard deviation vector and $\sigma^2$ is the diagonal vector of $\Sigma$. $\mathrm{LN}$, $\mathrm{fc}$ and $\mathrm{attn}$ represent the LayerNorm~\cite{ba2016layer}, linear layer and self-attention module~\cite{lin2017structured} respectively. 

\textbf{Multiple UA:}
In this paper, we propose to use multiple UA for multi-step uncertainty augmentation, which can obtain more different features from different distributions. With that, we can obtain more ranking information. 

Formally, suppose that there are $n$ UA modules, we need to learn $n$ probability distributions $\{\mathcal{N}(\mu_i, \Sigma_i)\}_{1,n}$. With the multiple UA, we can obtain a series of output features $\left(f_0, f_1, \cdots, f_i, \cdots,f_n\right)$, where $f_0 \in \{f_s,f_ t\}$ denotes the feature that has not been augmented with uncertainty augmenter, and $f_i$ is sampled from $\mathcal{N}\sim(\mu_i, \Sigma_i)$. In this feature sequences, $f_i$ is closer to the original feature $f_0$ than $f_j\ (i < j)$, i.e., $f_j$ has more uncertainty. Note that we can sample and obtain the source sequences 
\begin{equation}
   \boldsymbol{ \left(f_{s_0}, f_{s_1}, \cdots, f_{s_i}, \cdots,f_{s_n}\right) },
\end{equation}
and target sequences 
\begin{equation}
    \boldsymbol{ \left(f_{t_0}, f_{t_1}, \cdots, f_{t_i}, \cdots,f_{t_n}\right)},
\end{equation}
where there are $2n$ uncertainty augmenter modules for $f_s$ and $f_t$, respectively, as shown in Figure~\ref{fig:overview}. With the proposed multiple UA, we can obtain a list of samples from a triplet. 

However, sampling the feature from the distribution $\mathcal{N}\left( \mu, \Sigma \right)$ directly will prevent the gradient from back-propagating. To make the mean and standard deviation trainable, we use the reparameterization trick~\cite{kingma2014vae}:
\begin{equation}
    f_i=\mu_i+\sigma_i\epsilon,\quad\epsilon\sim\mathcal{N}(0,I).
\end{equation}
The architecture of UA is shown in Figure~\ref{fig:ua}.

\textbf{Remark 1:} We also use uncertainty learning in our proposed method. Different from the existing methods that evaluate the  uncertainty on a prediction, we use ranking-aware uncertainty to obtain more many-to-many correspondences, thus improving the retrieval ability. 

\subsection{Cross-Sample Uncertainty}
Further, we propose to exploit the uncertainty ranking information of other samples to establish many-to-many relationships through cross-sample uncertainty (CSU). We mind other positive samples from other triplets to reduce the conflict in the triplet optimization. 

Specifically, for a $i$-th source feature $f_s^i$ in the mini-batch, retrieving the target feature $f_t$ in the same batch and calculating the cosine similarity to get an ordered feature sequence $(f_t^1,f_t^2,\cdots,f_t^j)$. $f_t^i$ is closer to $f_s^i$ than $f_t^j$, i.e., the feature similarity of the source features and $f_t^i$ is larger.
We refine the contrastive loss to learn the cross-sample uncertainty loss:
\begin{equation}
    \begin{aligned}
        &{L}_{CS}(f_ s , f_ t) = \frac1B\sum_{i=1}^B-\log
        \frac{\exp\left(S\left(f_s^i,f_t^i\right)\right) + g(f_s^i, f_t)}
        {\sum_{j=1}^B\exp\left(S\left(f_s^i,f_t^j\right)\right)},\\
        &g(f_s, f_t) = \sum_{j=1}^B\exp\left(S\left(f_s,f_t^j\right)\right)\kappa\left(f_s,f_t^j\right),\\
        &\kappa\left(f_s,f_t\right) = \mathbb{I}\left(S\left(f_s,f_t\right) > cos(\theta)\right) \gamma, \\
    \end{aligned}
    \label{eq:L_CS}
\end{equation}
where $g(f_s, f_t)$ establishes the cross-sample correspondence. Specifically, $g(\cdot)$ retrieves the target samples in the batch that have a similarity to the source feature greater than $cos(\theta)$, then they are used as positive samples via the indicator function $\mathbb{I}(\cdot)$. $\theta$ is a threshold hyperparameter indicating the angle between the two features. $\gamma = 1 - \frac{current\_epoch}{total\_epoch}$ is employed to dynamically control the weights of the uncertainty samples.

Note that cross-sample uncertainty is orthogonal to in-sample uncertainty, hence we combine the two methods to improve the performance of TGIR. The final cross-sample uncertainty loss is as follows:
\begin{equation}
    \mathcal{L}_{CS} = \frac{1}{2n}\sum_{k=0}^n\sum_{m=0}^nL_{CS}(f_{s_k} , f_{t_m}).
\end{equation}

In the cross-sample uncertainty loss, we explore to mine the positive samples and learn the similarities with different levels of uncertainty.  

\subsection{Distribution Regularization}
As mentioned above, multiple UA learn multiple feature distributions to capture the rich semantic representations. To make the learned distributions meaningful, we align the feature distributions for each set of target and source distributions. It mitigates the problem of establishing incorrect many-to-many correspondences due to the multiple UA. We propose to  constrain UA to produce the same distribution for the target and source features. In our study, the 2-Wasserstein distance~\cite{Gulrajani2017wgan,Kim21vilt} was used to measure the distance between multivariate Gaussian distributions. 

For the source distribution $\mathcal{N}(\mu_s, \Sigma_s)$ and the target distribution $\mathcal{N}(\mu_t,\Sigma_t)$, the 2-Wasserstein distance can be defined as:
\begin{equation}
    \begin{aligned}
        D(\mu_s, \mu_t, \Sigma_{s}, \Sigma_{t}) &=\|\mu_s-\mu_t\|_{2}^{2}+\mathrm{Tr}((\Sigma_{s}^{1/2}-\Sigma_{t}^{1/2})^{2}) \\
        &=\|\mu_s-\mu_t\|_{2}^{2}+\|\sigma_{s}-\sigma_{t}\|_{2}^{2}.
    \end{aligned}
\end{equation}
Since distance and similarity are inversely proportional, for $n$ sets of target and source feature distributions, the distribution regularization loss is defined as:

\begin{equation}
    \mathcal{L}_{DR} = -\frac{1}{nB}\sum_{k=1}^n\sum_{i=1}^B\log\frac{\exp\left(-D(\mu_{s_k}^i, \mu_{t_k}^i, \Sigma_{s_k}^i, \Sigma_{t_k}^i) \right)}{\sum_{j=1}^B\exp\left(-D(\mu_{s_k}^i, \mu_{t_k}^j, \Sigma_{s_k}^i, \Sigma_{t_k}^j)\right)}.
\end{equation}

With the proposed cross-sample uncertainty loss and the distribution regularization loss,
the final loss function can be formulated as:
\begin{equation}
    \mathcal{L} = \frac{1}{2}(\mathcal{L}_{DR} + \mathcal{L}_{CS}).
\end{equation}

\subsection{Model Deployment}

\begin{figure}[htb]
    \centering
    \includegraphics[width=\linewidth]{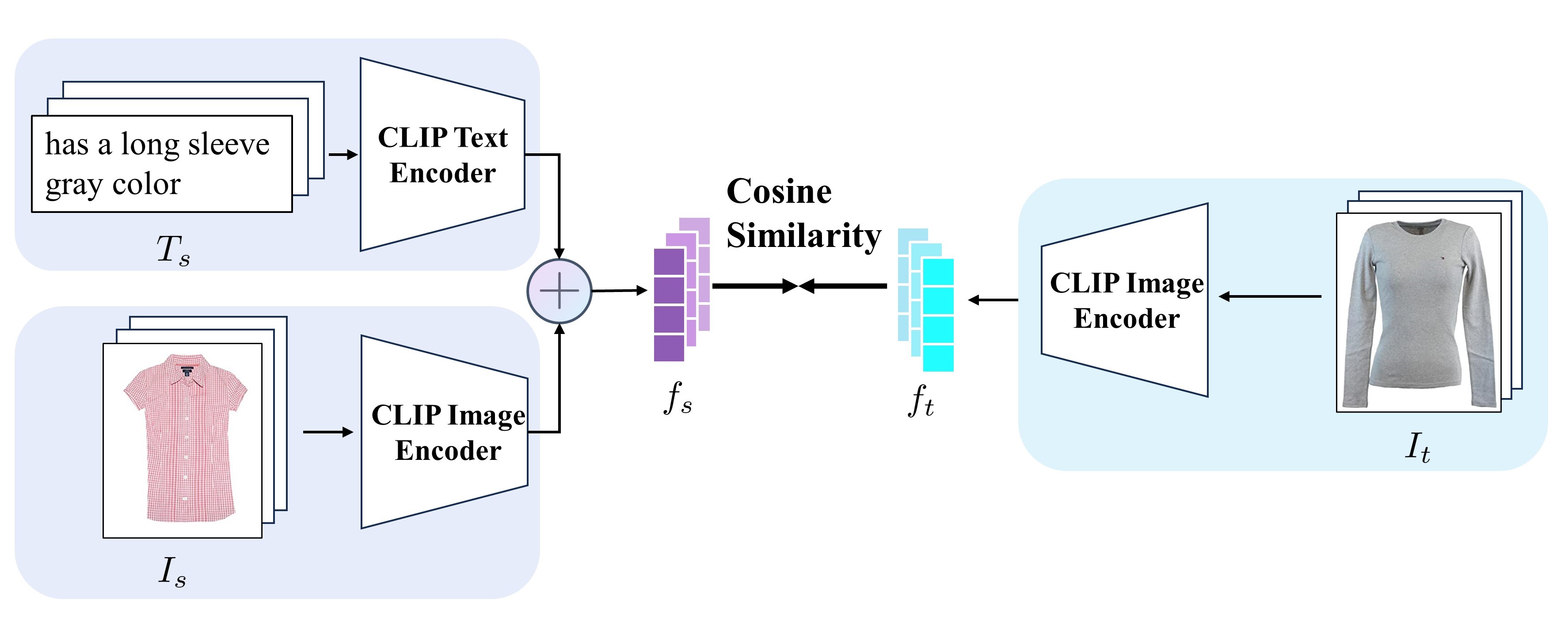}
    \caption{Pipeline of proposed model at test time.}
    \label{fig:test time}
\end{figure}

After the retrieval model is trained, as shown in Figure~\ref{fig:test time}, we first use the CLIP image encoder to compute the feature $f_t$ for all target images in the test database. Given a test pair with an image and a text, 
the input test image and test text first go through the CLIP to obtain the source features $f_s$. Then, we compute the cosine similarity between test feature and all target features, and the top $k$  images in the database are returned. Please note that we only use the features $f_s$ and $f_t$ when testing and the ranking-aware uncertainly module is removed. Thus, our method does not increase the retrieval times.  

\section{Experiments}
In this section, we present a comparative analysis of the performance of our proposed method against state-of-the-art approaches on two widely adopted datasets, FashionIQ and CIRR.

\begin{table*}[htb]
\centering
\small
    \begin{tabular}{l|c|cccccc|cc}
    \toprule
    \multirow{2}{*}{Methods} & \multirow{2}{*}{Visual Backbone} & \multicolumn{2}{c}{Dress} & \multicolumn{2}{c}{Shirt} & \multicolumn{2}{c}{Toptee} & \multicolumn{2}{|c}{Average}\\
    \cmidrule{3-4} \cmidrule{5-6} \cmidrule{7-8} \cmidrule{9-10}
    \multirow{2}{*}{} & \multirow{2}{*}{} &  R@10 & R@50 & R@10 & R@50 & R@10 & R@50 & R@10 & R@50\\
    \midrule
    MRN \cite{MRN} & ResNet-152 & 12.32 & 32.18 & 15.88 & 34.33 & 18.11 & 36.33 & 15.44 & 34.28 \\
    FiLM \cite{perez2018film} & ResNet-50 & 14.23 & 33.34 & 15.04 & 34.09 & 17.30 & 37.68 & 15.52 & 35.04 \\
    TIRG \cite{TIRG} & ResNet-17 & 14.87 & 34.66 & 18.26 & 37.89 & 19.08 & 39.62 & 17.40 & 37.39 \\
    CIRPLANT w/OSCAR \cite{liu2021image} & ResNet-152 & 17.53 & 38.81 & 17.45 & 40.41 & 21.64 & 45.38 & 18.87 & 41.53 \\
    VAL \cite{Chen_2020_CVPR} & ResNet-50 & 21.12 & 42.19 & 21.03 & 43.44 & 25.64 & 49.49 & 22.60 & 45.04 \\
    ARTEMIS \cite{delmas2022artemis} & ResNet-50 & 27.16 & 52.40 & 21.78 & 54.83 & 29.20 & 43.64 & 26.05 & 50.29 \\
    DCNet \cite{kim:2021:AAAI} & ResNet-50 & 28.95 & 56.07 & 23.95 & 47.30 & 30.44 & 58.29 & 27.78 & 53.89 \\
    CoSMo \cite{2021CoSMo} & ResNet-50 & 26.45 & 52.43 & 26.94 & 52.99 & 31.95 & 62.09 & 28.45 & 55.84 \\
    CLVC-Net \cite{CLVC-Net} & ResNet-50$\times$2 & 29.85 & 56.47 & 28.75 & 54.76 & 33.50 & 64.00 & 30.70 & 58.41 \\
    CLIP4Cir \cite{Baldrati2022combiner} & ResNet-50$\times$4 & \underline{31.63} & 56.67 & \underline{36.36} & 58.00 & \underline{38.19} & 62.42 & \underline{35.39} & 59.03 \\
    MGUR \cite{chen2022composed} & ResNet-50 & {{30.60}} & \underline{{57.46}} & {31.54} & \underline{58.29} & {{37.37}} & \underline{{68.41}} & {{33.17}} & \underline{{61.39}}\\
    \midrule
    Ours & ResNet-50 & \textbf{{34.80}} & \textbf{{60.22}} & \textbf{45.01} & \textbf{69.06} & \textbf{{47.68}} & \textbf{{74.85}} & \textbf{{42.50}} & \textbf{{68.04}}\\
    \bottomrule
    \end{tabular}
    \caption{Comparison results on FashionIQ validation set. The best performance is in bold, while the second-best is underlined. 
    Recall rate R@K, which signifies Recall@K (with higher values indicating superior performance). 
    The term ``Average'' refers to the mean value of corresponding R@K across sub-datasets.} 
    \label{tab:Recall rates on fashionIQ}
\end{table*}
\subsection{Implementation Details}
We employed Pytorch and performed all experiments on an NVIDIA RTX3090 graphics card. ResNet-50 and Transformer of the pre-trained model CLIP~\cite{Radford2021CLIP} were used as the image encoder and text encoder of our network backbone, respectively. The AdamW~\cite{Kingma2015adam} optimizer with an initial learning rate of 1e-6 was used for model training, which followed the training paradigm of the original CLIP and previous work~\cite{CLIP4Cir} on fine-tuning the CLIP model. In addition, the mini-batch size and epoch of training were set to 32 and 100, respectively. As for text and image preprocessing, we also followed CLIP's setting~\cite{Radford2021CLIP}. 
And the hyperparameters $\theta$ and $n$ in the proposed method were set to $45^\circ$ and 2, respectively. The code will be open-sourced to reproduce the experimental results of the method proposed in this paper.

\subsection{Datasets}
To verify the effectiveness of our proposed method, we conduct experiments on two real publicly available composed image retrieval datasets, including FashionIQ \cite{guo2019fashion} and CIRR \cite{liu2021image}. These datasets collect real feedback information from human users, which describes the user's modification intention.

\subsubsection{FashionIQ.}
FashionIQ \cite{guo2019fashion} is the pioneering fashion dataset that offers human-generated captions to discern similar pairs of garments, while also providing supplementary information in the form of authentic product descriptions and derived visual attribute labels for these images. Fashion IQ categories 77,684 fashion images from Amazon.com into three groups: Dress, Toptee, and Shirt. The dataset includes 18,000 triplets for training and 6,017 triplets for validation. Each triplet consists of a reference source image, a caption describing the modification intent, and a target image. The experimental setup adheres to the standard of previous work \cite{Chen_2020_CVPR, 2021CoSMo, CLIP4Cir, chen2022composed}.

\subsubsection{CIRR.}
The CIRR (Compose Image Retrieval on Real-life images) dataset \cite{liu2021image} broadens the horizons of compositional image retrieval to encompass open domains, requiring deep visual reasoning across rich image and language scenarios.  The dataset draws on 21,552 images from the renowned language reasoning dataset $\text{NLVR}^{2}$, featuring a varied and intricate spectrum of modification types, such as color, shape, position, number, size, and direction, as well as diverse and challenging images from open domains, such as animals, plants, vehicles, etc.  It comprises 36,554 triplets with the same format as FashionIQ and is partitioned into training, validation, and test sets in an 8:1:1 proportion. 

\subsubsection{Evaluation metric.}
Following the previous work \cite{liu2021image, CLIP4Cir}, we use Recall within Top-K (Recall@K) as the composed image retrieval evaluation metric, which measures the percentage of at least one correctly retrieved image appearing in the top K retrieved items. In addition, thanks to CIRR's unique dataset involvement, we additionally report $\text{Recall}_{\text{subset}}@K$ ($\text{R}_{\text{subset}}@K$), which only considers images in the query subset. $\text{R}_{\text{subset}}@K$ is not affected by false negative samples and helps analyze the reasoning performance of models that capture fine-grained image-text modifications by selecting a batch of negative samples with high visual similarity. We also report the mean of R@5 and R$_\text{subset}@1$ as the overall performance of our model on CIRR~\cite{liu2021image}.

\begin{table*}[]
  \centering 
  \resizebox{\linewidth}{!}{
  \begin{tabular}{lcccccccc} 
  \toprule
  \multicolumn{1}{c}{}        & \multicolumn{4}{c}{Recall$@K$ }               & \multicolumn{3}{c}{Recall$_{\text{Subset}}@K$ }   & \multirow{2}{*}{(R$@5$ $+$ R$_{\text{Subset}}@1$)$/2$}       \\
  \cmidrule(lr){2-5}
  \cmidrule(lr){6-8}
  \multicolumn{1}{l}{Methods} & $K=1$           & $K=5$           & $K=10$ & $K=50$          & $K=1$           & $K=2$           & $K=3$            &  \\ 
  \midrule
    
    TIRG~\cite{TIRG}             & 14.61  & 48.37  & 64.08  & {90.03} &  22.67  & 44.97  & 65.14  & 35.52 \\ 
    TIRG$+$LastConv~\cite{TIRG}             & 11.04  & 35.68  & 51.27  & 83.29 &  23.82  & 45.65  & 64.55  & 29.75 \\ 
    MAAF~\cite{dodds2020modality_maaf}         & 10.31     & 33.03     & 48.30    & 80.06 &  21.05    & 41.81    & 61.60  & 27.04 \\
    MAAF$+$BERT~\cite{dodds2020modality_maaf}  & 10.12     & 33.10     & 48.01 & 80.57    & 22.04    & 42.41    & 62.14   & 27.57 \\
    MAAF$-$IT~\cite{dodds2020modality_maaf}    & 9.90      & 32.86     & 48.83 & 80.27    & 21.17    & 42.04    & 60.91   & 27.02 \\
    MAAF$-$RP~\cite{dodds2020modality_maaf}    & 10.22     & 33.32     & 48.68 & 81.84    & 21.41    & 42.17    & 61.60   & 27.37 \\
    
    CIRPLANT \cite{liu2021image}   & {15.18}  & 43.36  & 60.48  & 87.64  & 33.81   & 56.99  & 75.40  & {38.59} \\ 
    CIRPLANT w/OSCAR \cite{liu2021image} & {19.55}  & {52.55}  & {68.39}  & {92.38} &  {39.20}  & {63.03}  & {79.49}  & {45.88} \\ 
    CLIP4Cir \cite{Baldrati2022combiner} & \textbf{33.59} & \underline{65.35} &  \underline{77.35} & \underline{95.21} & \textbf{62.39} & \textbf{81.81} & \textbf{92.02} & \underline{63.87} \\
    \midrule
    Ours & \underline{32.24} & \textbf{66.63} & \textbf{79.23} & \textbf{96.43} & \underline{61.25} & \underline{81.33} & \textbf{92.02} & \textbf{63.94} \\
  \bottomrule
  \end{tabular}}
  \caption{Comparison results on CIRR official test set. (R$@5$ $+$ R$_{\text{Subset}}@1$)$/2$ represent the overall performance of the method. The best performance is in bold and the second-best is underlined.}
  \label{tab:Recall rates on CIRR}
\end{table*}

\subsection{Comparison with State-of-the-Art Methods}
To demonstrate the superiority of proposed method, we compare the results of the proposed method with state-of-the-art (SOTA) models on the publicly available FashionIQ and CIRR dataset. 

\subsubsection{Comparison on FashionIQ.}
In the baseline models, CLIP4Cir~\cite{Baldrati2022combiner} is the SOTA model on R@10. It uses pre-trained CLIP 4 $\times$ ResNet-50 and Transformer as encoders to extract image and text features respectively, and uses contrast learning to train a merger network to get a convex combination of text and image features. MGUR~\cite{chen2022composed} is the SOTA model on R@50. It introduces Gaussian noise in the feature space and uses uncertainty regularisation to adaptively match object according to the range of noise fluctuations. 

Table~\ref{tab:Recall rates on fashionIQ} shows the retrieval performance on the FashionIQ validation set. 
We can observe that our proposed method greatly outperforms all SOTA models, which validates the motivation of uncertainty-aware ranking to mine more potential candidates by establishing many-to-many relationships. Specifically, our method significantly outperforms R@10 for CLIP4Cir and R@50 for MGUR by margins of 7.11\% and 6.65\%, respectively. The average of R@10 is 42.50\% compared to 33.17\% of MGUR, which indicates the benefits of the proposed ranking-aware uncertainty over the data enhancement method that also uses Gaussian distribution.

\subsubsection{Comparison on CIRR.}

As listed in Table \ref{tab:Recall rates on CIRR}, we provide the results on the CIRR test set obtained through the official evaluation server. CLIP4Cir~\cite{Baldrati2022combiner} is the SOTA model on the CIRR dataset.
Moreover, our approach achieves the SOTA overall performance (63.94\% (R$@5$ $+$ R$_{\text{Subset}}@1$)$/2$). Specifically, the proposed model outperforms the previous best model~\cite{Baldrati2022combiner} in Recall@5, Recall@10, and Recall@50 metrics, indicating that many-to-many correspondence can facilitate the model to capture coarse-grained modifications between similar images. Note that CLIP4Cir uses a scaled 4 $\times$ ResNet-50 that follows the EfficientNet style~\cite{Tan2019icml} as a visual coder, with a much larger number of parameters than the proposed method. Furthermore, our method significantly outperforms the second SOTA method CIRPLANT~\cite{liu2021image} using ResNet152 as a visual backbone with an overall performance of 18.06\%. Overall, the fact that our model achieves such competitive results with fewer parameters illustrates that our approach is more effective. B, the images of this dataset were grouped into multiple subsets of six images that were semantically and visually similar, and relevant captions were collected to describe the differences between two images within the same subset.

\subsection{Ablation Studies}

\begin{threeparttable}
    \centering
    \resizebox{\linewidth}{!}{
    \begin{tabular}{l|ccccc}
    \toprule
    \multirow{2}{*}{Method} & \multicolumn{4}{c}{Average}\\
    \cmidrule{2-5}
    \multirow{2}{*}{} & R@1 & R@5 & R@10 & R@50\\
    \toprule
    Baseline & 13.14 & 29.46 & 38.19 & 62.78 \\
    $+$ CSU & 13.56 & 31.55 & 41.71 & \underline{68.07}\\ 
    $+$ ISU & 14.00 & 31.71 & \underline{42.11} & 67.61\\
    $+$ ISU $+$ CSU & 13.50 & 31.57 & 42.05 & \textbf{68.39}\\
    $+$ ISU $+$ CSU $+$ DR & \textbf{14.57} & \textbf{32.00} & \textbf{42.50} & 68.04\\
    \bottomrule
    \end{tabular}
    }
    \caption{Ablation study on FashionIQ. 
    }
    \label{tab:Ablation}
\end{threeparttable}

We perform an ablation study to verify the effectiveness of each module in proposed model. We first set up a baseline model without ranking-aware uncertainty, i.e., with only CLIP image and text encoders as shown in Figure~\ref{fig:ua}. The hyper-parameter setting of the baseline model is retained the same as the proposed method, except that the training loss is replaced by $\mathcal{L}_{CL}$. The specific variants of our
model is described as follows:
\begin{itemize}
    \item  CSU: To study the effect of cross-sample uncertainty separately, we add the cross-sample uncertainty method to the baseline model.
    \item ISU: To study the effect of in-sample uncertainty separately, we add the in-sample uncertainty method to the baseline model.
    \item  ISU $+$ CSU: To check the effect of the combination of ISU and CSU, we use both methods on Baseline.
    \item ISU $+$ CSU + DR (our proposed method): To investigate whether distribution regularization can mitigate the degradation of fine-grained model retrieval due to excessive uncertainty.
\end{itemize}

As listed in Table \ref{tab:Ablation}, we obtain three observations as follows: (1) Both ISU and CSU can bring significant improvement to the retrieval performance of the model when used individually. Where CSU is better than ISU in terms of R@50 with an improvement of 5.29\% compared to baseline. While ISU achieves 3.92\% improvement in R@10. (2) The simultaneous use of ISU and CSU reduces the model's fine-grained retrieval ability, but the model's coarse-grained retrieval ability is best (68.39\% R@50). (3) DR effectively mitigates the problem of incorrect correspondence arising from the simultaneous use of ISU and CSU by aligning feature distributions, which improves the model's fine-grained retrieval capability (R@1, R@5 and R@10 improved by 0.93\%, 0.43\% and 0.45\% respectively).

\subsection{Hyper-parameter Tuning}

\begin{table}[]
    \centering
   \begin{tabular}{l|ccccc}
    \toprule
    \multirow{2}{*}{$\theta$} & \multicolumn{4}{c}{Average}\\
    \cmidrule{2-5}
    \multirow{2}{*}{} & R@1 & R@5 & R@10 & R@50\\
    \toprule
    $75^{\circ}$ & \underline{14.41} & \underline{31.75} & 41.90 & \textbf{68.26} \\
    $60^{\circ}$ & 14.14 & 31.70 & 42.02 & 67.78 \\
    $45^{\circ}$ & \textbf{14.57} & \textbf{32.00} & \underline{42.50} & \underline{68.04} \\
    $30^{\circ}$ & 14.08 & 31.93 & \textbf{42.60} & 67.68 \\
    \bottomrule
    \end{tabular}
    \caption{The tuning of hyper-parameter $\theta$ on FashionIQ dataset.}
    \label{tab:hyper-parameter theta}
\end{table}

\begin{table}[]
    \centering
     \begin{tabular}{l|ccccc}
    \toprule
    \multirow{2}{*}{$n$} & \multicolumn{4}{c}{Average}\\
    \cmidrule{2-5}
    \multirow{2}{*}{} & R@1 & R@5 & R@10 & R@50 \\
    \toprule
    $1$ & \textbf{14.69} & \textbf{32.02} & \underline{42.06} & 68.01 \\
    $2$ & \underline{14.57} & \underline{32.00} & \textbf{42.50} & \underline{68.04} \\
    $3$ & 13.68 & 30.95 & 41.90 & \textbf{68.06} \\
    \bottomrule
    \end{tabular}
    \caption{The tuning of hyper-parameter $n$ on FashionIQ dataset.}
    \label{tab:hyper-parameter n}
\end{table}

As mentioned above, our proposed method contains hyper-parameters $\theta$ and $n$. Specifically, $\theta$ denotes the angle between two vectors, and according to Eq. \ref{eq:L_CS}, when the angle between two vectors is less than $\theta$, it is considered as a potential matching sample. And $n$ denotes the number of UA on the $f_s$ or $f_t$ side. As listed in Table~\ref{tab:hyper-parameter theta}, the overall performance is best when $\theta=45^\circ$. In addition, the model remains robust to $\theta$ changes and achieves reasonable performance.

As listed in Table \ref{tab:hyper-parameter n}, our model achieves the best overall performance at $n=2$. When $n=1$ introduces less uncertainty, thus the model has a stronger fine-grained retrieval capability. Whereas, too much uncertainty is introduced at $n=3$, which may establish wrong many-to-many relationships, thus affecting the model's fine-grained matching ability. In addition, R@50 fluctuates less when $n$ varies, indicating that our model's coarse-grained retrieval ability is robust.

\section{Conclusion}
In this paper, we proposed a ranking-aware uncertainty method for text-guided image retrieval. It provides an early attempt to solve the problem that existing triplet optimization methods cannot account for the many-to-many correspondences in feedback languages and images due to semantic diversity. We proposed in-sample uncertainty to expand one-to-one triples into many-to-many correspondences. And cross-sample uncertainty is used to mine the possible correspondences between different triples. Then, distribution regularization was proposed to align the target and source feature distributions. An empirical evaluation of extensive experiments showed that the proposed method has better performance than state-of-the-art baselines.

\bibliography{references}
\end{document}